\documentclass[10pt,conference]{IEEEtran}
\usepackage{header}
\usepackage{subcaption}

\author{
    \IEEEauthorblockN{Cleison Correia de Amorim}
    \IEEEauthorblockA{
        \textit{Centro de Informática} \\
        \textit{Universidade Federal de Pernambuco} \\
        50.740-560, Recife, PE, Brazil \\
        cca5@cin.ufpe.br
    }
    \and
    \IEEEauthorblockN{Cleber Zanchettin}
    \IEEEauthorblockA{
        \textit{Centro de Informática} \\
        \textit{Universidade Federal de Pernambuco} \\
        50.740-560, Recife, PE, Brazil \\
        cz@cin.ufpe.br
    }
}

\begin{document}
    \title{\datasetTD{} and \datasetPhono{}: Two Novel Datasets for the American Sign Language}

    \maketitle
    
    \begin{abstract}
    Sign language is an essential resource enabling access to communication and proper socioemotional development for individuals suffering from disabling hearing loss. As this population is expected to reach 700 million by 2050, the importance of the language becomes even more essential as it plays a critical role to ensure the inclusion of such individuals in society. The Sign Language Recognition field aims to bridge the gap between users and non-users of sign languages. However, the scarcity in quantity and quality of datasets is one of the main challenges limiting the exploration of novel approaches that could lead to significant advancements in this research area. Thus, this paper contributes by introducing two new datasets for the American Sign Language: the first is composed of the three-dimensional representation of the signers and, the second, by an unprecedented linguistics-based representation containing a set of phonological attributes of the signs.
    \end{abstract}
    
    
    \section{Introduction} 
\label{sec:introduction}

According to projections of the World Health Organization (WHO), the number of individuals with disabling hearing loss\footnote{Disabling hearing loss refers to the loss greater than 35 decibels (dB) in the better hearing ear~\cite{who2021report}.} is close to 450 million and, as the world's population continues to grow, it is anticipated that by 2050 this number will exceed 700 million people. This means that nearly 1 in 14 individuals (at least 7\%) will require hearing care by 2050~\cite{who2021report}. 

However, despite effective interventions such as medical and surgical management, hearing aids, cochlear implants, and rehabilitative therapy, the vast majority of those in need do not have access to them. In addition, most people with hearing loss live in low-income settings where human resources and services for ear and hearing care are not commonly accessible~\cite{who2021report}. 

Due to those limitations, sign language emerges as one accessible alternative that can ensure to those individuals access to communication, as well as the proper cognitive and socioemotional development of deaf infants and children~\cite{who2021report}. It is essential to support research that broad the awareness and access to this language across society.

The use of computational approaches to help the learning, communication, translation, and interaction using sign language is essential to include this population in society. In the machine learning research area, Sign Language Recognition (SLR) refers to a collaborative field that involves pattern matching, computer vision, natural language processing, and linguistics \cite{wadhawan2021review} to enable sign language users to communicate with spoken languages users. However, while speech recognition systems have advanced to the point of being commercially available, SLR systems are still in very early stages. Therefore, commercial translation services for sign languages are human-based, require specialized personnel, and are often expensive~\cite{cooper2011slr}.

Among the main challenges involved, there is the scarcity of approaches structured around the linguistics of the sign language, the limitation in quantity and quality of public datasets that could lead to significant advancements of the SLR techniques, and the multi-channel nature of the language -- which convey meaning through many modes at once~\cite{cooper2011slr}.

By linguistics, we understand the scientific study of the natural and human languages that aims at unraveling independent and universal principles of logic and information that determine the language. It seeks to answer essential questions such as: "What is the nature of human language?", "How is communication built?", "What are the principles that determine the ability of human beings to produce and understand language?" \cite{quadros2007lingua}.

Linguistics is divided into areas that address different aspects of the language, among which the phonology studies the sounds of a language from a functional perspective, as elements that integrate a linguistic system \cite{quadros2007lingua}. 

Although the notion of phonology is traditionally based on sound systems, phonology also includes the equivalent component of the grammar in sign languages because it is tied to the grammatical organization and not to the particular content \cite{brentari2018slphonology}. Thus, when it comes to sign languages, its first task is to determine the minimum units that make up the signs. Then, the second task is to establish what are the possible patterns of combination between these units, as well as the possible variations in the phonological environment \cite{quadros2007lingua}.

Stokoe \cite{stokoe1960slstructure} proposed in 1960 the first structural linguistic schema to analyze the formation of signs, as well as their decomposition into main aspects or parameters that do not carry meaning in isolation. Since then, other phonologists have extended the representation to the one widely acknowledged in the sign language literature, which is composed by the manual (handshape, location, movement, and orientation) and non-manual parameters (such as the expressions of the face, eyes, head or trunk), as detailed below \cite{quadros2007lingua,brentari2018slphonology,sandler2012phonological}:

\begin{itemize}
\item Handshape (or hand configuration): the configuration assumed by the hand while producing the sign, with one or more selected fingers in a particular position -- extended, closed, curved, or bent.
\item Orientation (or palm orientation): the direction the palm points while producing the sign.
\item Location (or point of articulation): the area on the body, or in the articulation space, at which or near which the sign is articulated.
\item Movement: complex parameter that may involve a wide range of modes and directions, such as the internal movements of the hands, wrist, and directional movements of the hands in the space.
\item Non-manual expressions: movements performed by the face, eyes, head, and trunk. If we consider a more granular level, this might also include mouth movements, eyebrows, cheeks, shoulders, and others. They perform two essential functions: 1) marking syntactic constructions (interrogative sentences, relative clauses, topicalizations, agreement, and focus) and 2) differentiating lexical components (specific references, pronominal references, negative particles, adverbs, degree, or aspect).
\end{itemize}

In order to contribute to overcoming the challenges presented above and enable the adoption of techniques that may lead to new advancements in the SLR field, this work introduces two datasets derived from one of the most relevant sign language datasets -- the American Sign Language Lexicon Video Dataset (ASLLVD)~\cite{athitsos-asllvd-2008,neidle-2012}:

\begin{enumerate}
\item \datasetTD{}: provides a three-dimensional representation of the signers' body joints, allowing the complete observation of the signs articulation, frame-by-frame. Today, most of the datasets are limited to images and videos captured in two-dimensional space.
\item \datasetPhono{}: provides a linguistics-based representation that describes the signs in terms of phonological attributes of the American Sign Language (ASL), frame-by-frame along with the signs articulation. As stated above, linguistics-based approaches are scarce and this dataset is an unprecedented important resource to support researches in this direction.
\end{enumerate}

There was a set of computational challenges involved in designing and building the proposed datasets, among which we can enumerate the following:
\begin{itemize}
\item Establish an approach to reconstruct the 3D space from the 2D video sequences in the ASLLVD, as well as to handle missing or low-quality samples found.
\item Identify from ASL phonology the initial set of attributes that are able to track variations in the most significant body parts during the articulation of signs and that can be computationally represented.
\item Identify the mathematical and anthropometric approaches that enable the computation and representation of the dataset's attributes.
\item Computing more than 9,000 samples for each of the datasets, which required expensive resources such as 40 hours of distributed CPU and GPU processing and more than 1 TB of data.
\end{itemize}

    \section{Related work}
\label{sec:related-work}

According to~\cite{bragg2020exploring}, while powerful machine learning algorithms require large amounts of data, many application domains are still data-scarce. In particular, collecting sufficient data from small, underserved populations to build systems serving those groups is difficult because the pool of potential contributors is significantly reduced. 

The problem of collecting data from small groups is exemplified in sign language data collection. Although there are about 150 registered sign languages, their total number of users comprises only 25 million individuals, less than 0.32\% of the world population. As a result, sign language video corpora are typically small~\cite{ethnologue2021,bragg2020exploring}.

In addition, most of the relevant datasets are limited to providing two-dimensional images or videos captured from a single perspective of the individuals~\cite{quiroga2021slrdatasets,ms-asl-2019,lsa64-2016,rwth-phoenix-2015,signum-2007}. This imposes limitations on the range of techniques adopted to explore and build tools for the language. On the other hand, estimating different representations from single 2D images can be an expensive and challenging task.

\begin{figure}
    \centering
    \begin{subfigure}{0.32\linewidth}
        \centering
        \includegraphics[width=\linewidth]{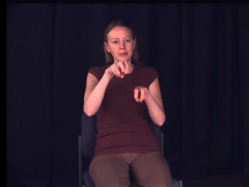}
        \caption{}
        \label{subfig:asllvd-example-front}
    \end{subfigure} %
    \begin{subfigure}{0.32\linewidth}
        \centering
        \includegraphics[width=\linewidth]{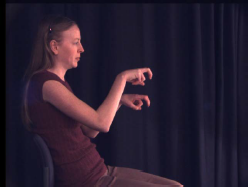}
        \caption{}
        \label{subfig:asllvd-example-side}
    \end{subfigure} %
    \begin{subfigure}{0.32\linewidth}
      \centering
      \includegraphics[width=\linewidth]{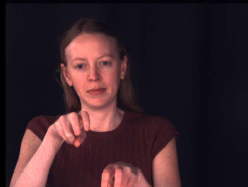}
      \caption{}
      \label{subfig:asllvd-example-close}
    \end{subfigure}%
    \caption{
       Sample of three synchronized perspectives of the same sign in the ASLLVD dataset: frontal view~(\subref{subfig:asllvd-example-front}), side view~(\subref{subfig:asllvd-example-side}), and face view~(\subref{subfig:asllvd-example-close}). Depending on the camera's point-of-view, different information and features can help train automatic classifiers to recognize the signs. The sample frame refers to the sign MERRY-GO-ROUND~\cite[p. 2]{athitsos-asllvd-2008}.
    }
    \label{fig:asllvd-example}
\end{figure}

The ASLLVD consists of a broad public dataset\footnote{Available at \url{http://www.bu.edu/asllrp/av/dai-asllvd.html}} of the American Sign Language (ASL) containing approximately 2,745 signs represented in about 9,763 annotated video sequences articulated by native individuals. Unlike what happens to the datasets mentioned above, it is composed of video sequences captured through four synchronized cameras: one side view of the signer, one close-up of the head region, one half-speed high-resolution front view, and another full resolution front view, as illustrated in the~\autoref{fig:asllvd-example}~\cite{athitsos-asllvd-2008,neidle-2012}. 

These different perspectives make it possible to extract a more significant number of details about the articulation of the signs and derive new representations without the need to capture the sequences again. Previously, we accomplished this in \cite{st-gcn-sl-2019} by introducing a two-dimensional representation for the skeletons of the signers. In the current work, we will extend this approach and introduce two new representations.

    \section{\datasetTD{}}
\label{sec:dataset-3d}

The \datasetTD{} introduces a representation based on mapping into the three-dimensional space the coordinates of the signers in the ASLLVD dataset. This enables a more accurate observation of the body parts and the signs articulation, allowing researchers to better understand the language and explore other approaches to the SLR field.

\begin{figure}
    \centering
    \begin{subfigure}{0.22\linewidth}
        \centering
        \includegraphics[height=3cm]{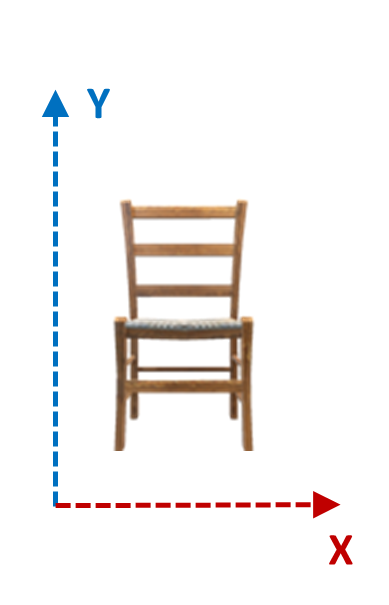}
        \caption{}
        \label{subfig:our-strategy-3d-front}
    \end{subfigure} %
    \begin{subfigure}{0.25\linewidth}
        \centering
        \includegraphics[height=3cm]{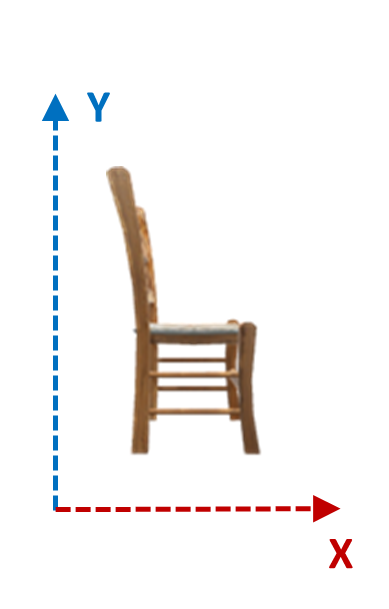}
        \caption{}
        \label{subfig:our-strategy-3d-side}
    \end{subfigure} %
    \begin{subfigure}{0.46\linewidth}
      \centering
      \includegraphics[height=3cm]{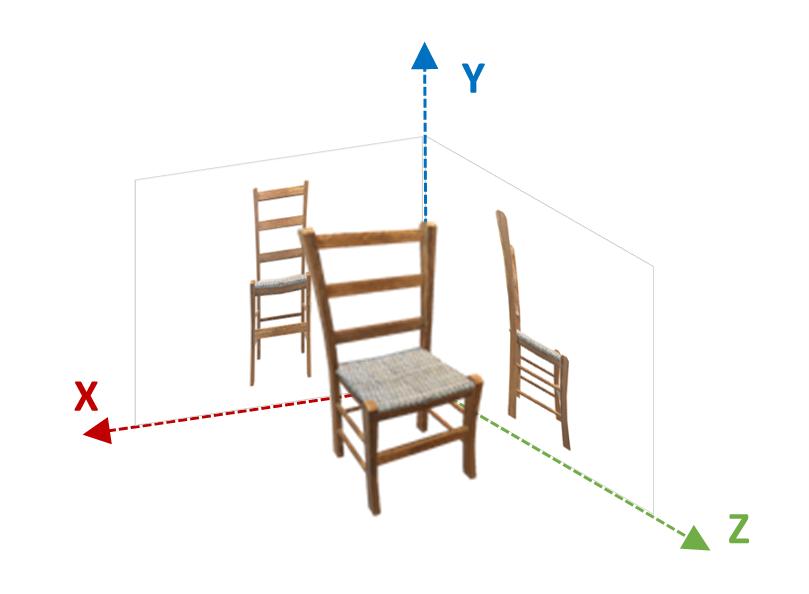}
      \caption{}
      \label{subfig:our-strategy-3d-persp}
    \end{subfigure}
    \caption{
       Strategy adopted for representing the signs into the three-dimensional space: the two-dimensional views for the front~(\subref{subfig:our-strategy-3d-front}) and side~(\subref{subfig:our-strategy-3d-side}) perspectives are positioned perpendicularly to reconstruct a three-dimensional view of the object~(\subref{subfig:our-strategy-3d-persp}).
    }
    \label{fig:our-strategy-3d}
\end{figure}

The strategy adopted for representing the signs into the three-dimensional space consists of first selecting two bi-dimensional perspectives arranged perpendicularly to each other: the frontal and the side view. Given two cameras positioned at a 90º angle, capturing the same target, the frontal perspective will allow the observation through the \(x\) and \(y\) axes, and the side view will observe through the depth dimension (or \(z\) axis) (see \autoref{fig:our-strategy-3d}).

To create the \datasetTD{}, we adopted the same steps presented in~\cite{st-gcn-sl-2019}, which comprises retrieving the video sequences, segmenting the signs, estimating the skeletons, and normalizing the output. However, several adaptations were required at each step to accommodate the strategy above, as described in the following sections.

\subsection{Obtaining original samples}
\label{subsec:obt-original-samples}

We started by fetching the samples from the ASLLVD, but now considering both the frontal and side cameras. There are two formats originally available for the cameras: \textit{mov} (compressed videos, usually smaller and easier to process) and \textit{vid} (raw videos, longer to download and process).

While downloading the samples, we identified that for part of them, the cameras were available in both the \textit{mov} and \textit{vid} formats. For other cases, each camera was available in a distinct format. However, in the worst cases, one of the cameras was just missing or corrupted, forcing us to discard a few samples.

Dealing with the problem above added additional complexity to our work. However, we have successfully overcome it once we missed only 0.16\% of the original samples, yielding a total of 9,747 samples.

\subsection{Segmenting signs}
\label{subsec:segment-signs}

While extracting the sub-sequences for each sign from the original videos, we reduced the frame rate from 60 to 3 FPS. This was made because, given the conditions in which the signers were captured in the dataset,  we assume they could not do more than three distinct movements in a single second.

In addition, this allowed us to simplify the operation by skipping the extraction and processing of keypoints that captured few significant changes.

\subsection{Estimating 3D skeletons}
\label{subsec:estimate-skeletons}

\begin{figure}
    \centering
    \begin{subfigure}{0.35\linewidth}
        \centering
        \includegraphics[width=1.\linewidth]{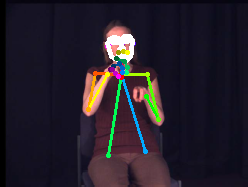}
        \caption{}
        \label{subfig:front-side-persp-skeletons-front}
    \end{subfigure} %
    \begin{subfigure}{0.35\linewidth}
        \centering
        \includegraphics[width=1.\linewidth]{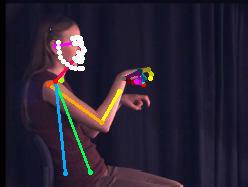}
        \caption{}
        \label{subfig:front-side-persp-skeletons-side}
    \end{subfigure} %
    \begin{subfigure}{0.65\linewidth}
      \centering
      \includegraphics[width=\linewidth]{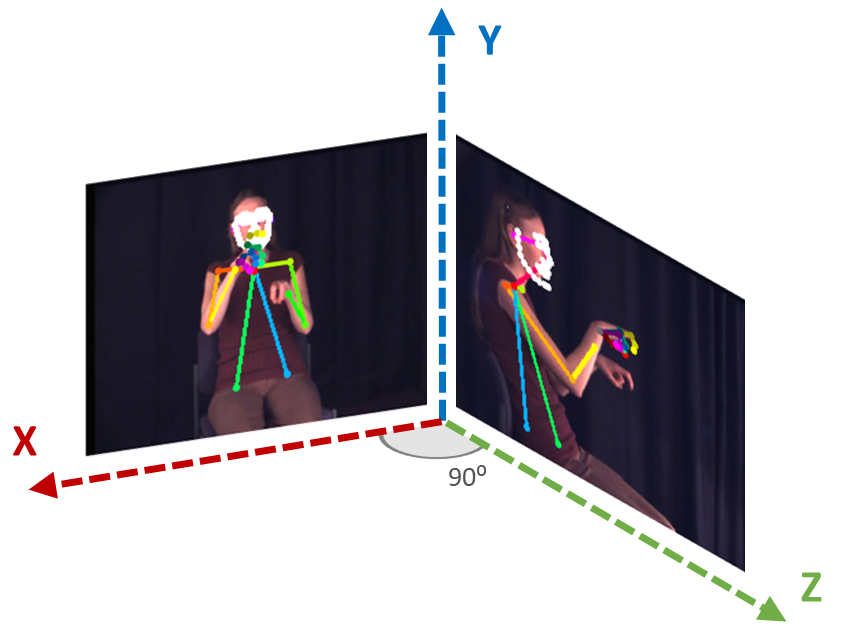}
      \caption{}
      \label{subfig:front-side-persp-skeletons-persp}
    \end{subfigure} %
    \caption{
        The \(x\) and \(y\) coordinates of our final 3D skeleton are taken from the keypoints estimated for the frontal perspective~(\subref{subfig:front-side-persp-skeletons-front}). The \(z\) coordinate is originated from the \(x\) axis of the side skeleton~(\subref{subfig:front-side-persp-skeletons-side}). When placed together, they are visualized like in~(\subref{subfig:front-side-persp-skeletons-persp}).
    }
    \label{fig:front-side-persp-skeletons}
\end{figure}

Next, we estimated the skeletons for the signers by using the OpenPose~\cite{openpose-cao-2019,openpose-hand-face-simon-2017}. This was made separately for both the frontal and side cameras, which yielded two bi-dimensional skeletons of the same individual:

\begin{enumerate}
\item The \textbf{frontal skeleton}  (\autoref{subfig:front-side-persp-skeletons-front}) contains the coordinates for body, hands, and head plotted over the \(x\) and \(y\) axes. If we consider building a 3D skeleton observed through the frontal perspective, these coordinates are a very good starting point to anchor the individual into the three-dimensional space. However, at this point, the coordinates in our partial 3D skeleton are still flat, lacking details about the depth dimension.
\item The \textbf{side skeleton} (\autoref{subfig:front-side-persp-skeletons-side}) also contains a set of \(x\) and \(y\) coordinates. However, suppose we place it perpendicularly to the previous view (as in the~\autoref{subfig:front-side-persp-skeletons-persp}). In that case, we will observe that although the \(y\) axis provides the same data as in the frontal skeleton, the \(x\) axis will add the depth dimension that was just missing. So, from now on, we will take the \(x\) axis of the side skeleton as being the \(z\) axis (depth axis) of our final 3D skeleton.
\end{enumerate}

\subsection{Normalizing 3D skeletons}
\label{subsec:normalize-skeletons}

Then, we normalized the 3D skeletons to remove variations arising from camera placement and differences in individuals' bodies. This is necessary because the ASLLVD dataset was captured over multiple sections and different signers' contribution.

\begin{figure}
    \centering
    \includegraphics[width=0.4\linewidth]{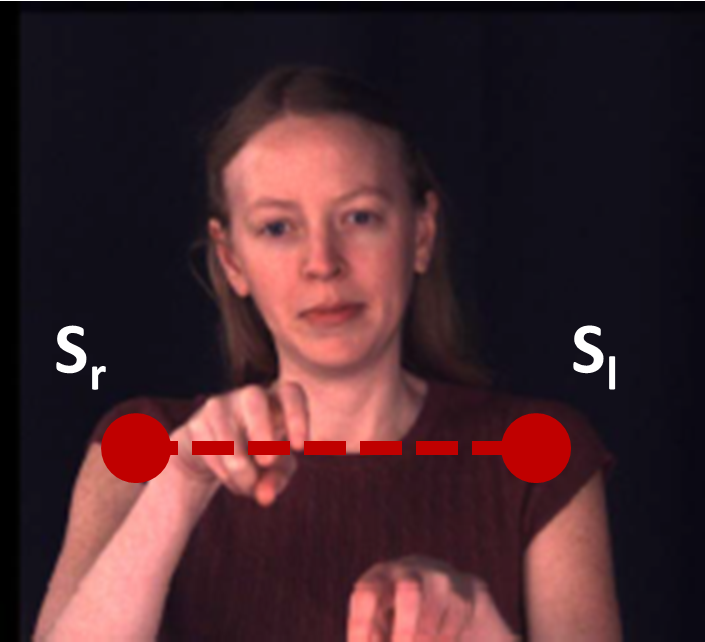}
    \caption{The width between the signer's shoulders was used as the reference to normalize the coordinates in the 3D skeletons.}
    \label{fig:shoulders-width}
\end{figure}

For this, we adopted as a reference the width between the shoulders of the signers (see~\autoref{fig:shoulders-width}), which was inspired in the anthropometric measure \textit{biacromial diameter}~\cite{stoudt1970skinfolds}. Here, the shoulders width \(W_{shoulders}\) was calculated by computing the Euclidean distance \(d\)~\cite{anton2013algebra} between the coordinates of the left \(S_{l}\) and right \(S_{r}\) shoulders, as follows:

\begin{equation}
    \label{eqn:shoulders-width}
    W_{shoulders} = d\left(S_{l}, S_{r}\right)
\end{equation}

Once the \(W_{shoulders}\) was calculated, it was possible to transform each keypoint \(K\) into the respective normalized keypoint \(K_{norm}\):

\begin{equation}
    \label{eqn:normalized-keypoint}
    K_{norm} = \frac{K}{W_{shoulders}}
\end{equation}

\begin{figure}
    \centering
    \includegraphics[width=0.99\linewidth]{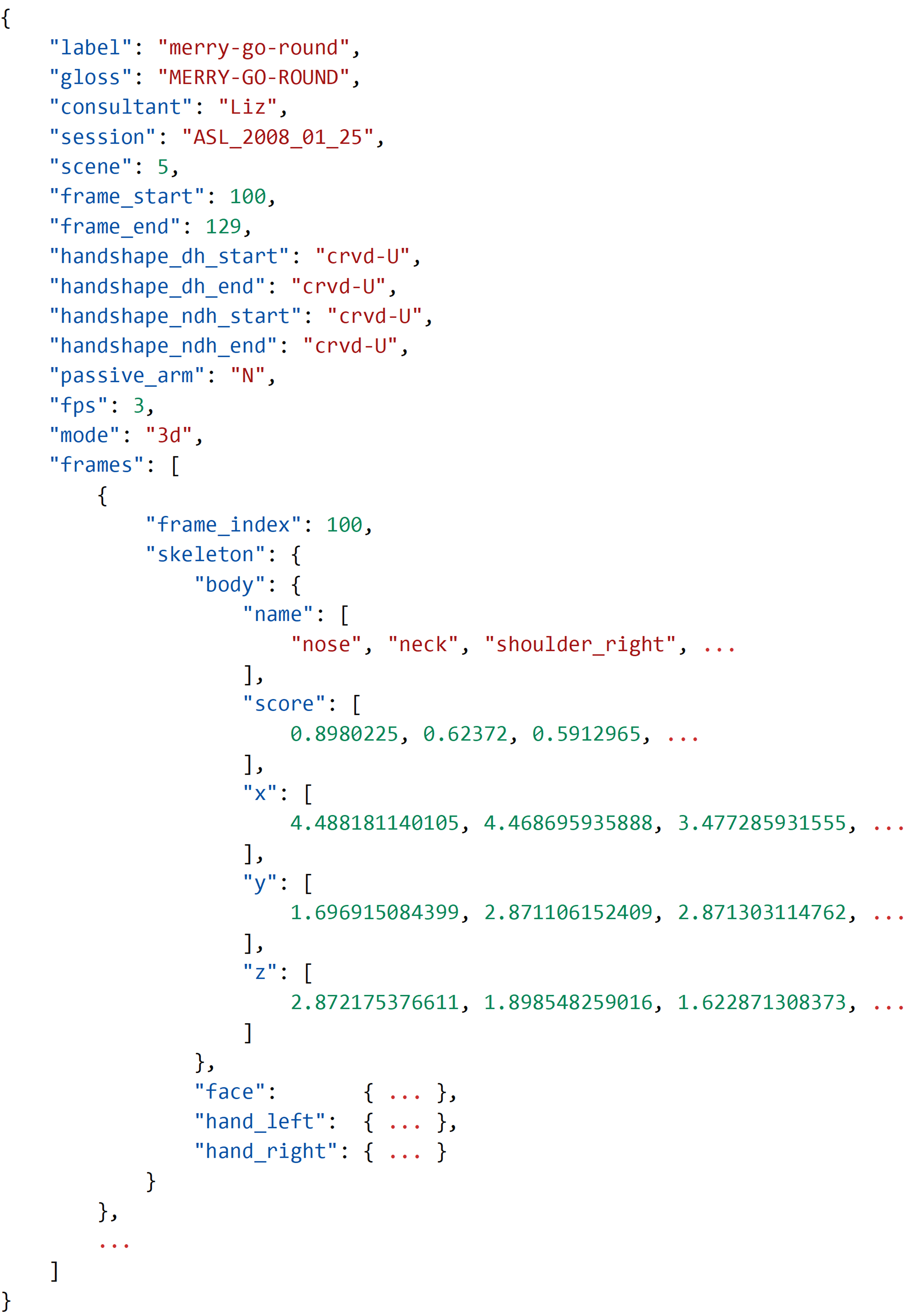} 
    \caption{Example of structure and content provided for the samples in the \datasetTD{} dataset.}
    \label{fig:sample-json-datasetTD}
\end{figure}

\autoref{fig:sample-json-datasetTD} illustrates an example of the information found for the samples in the \datasetTD{} dataset. The properties at the beginning identify the sign label and the session, scene, consultant, and frames in which it was originally recorded. There is also a property describing the sequence of skeletons estimated for the sign. In this property, for every frame, there are the keypoints grouped into the body, face, left hand, and right hand nodes, which contain respectively:

\begin{itemize}
\item \textbf{name}: the name of the body part to which the keypoint at that index refers. For example, the first index contains the name "nose", which means that the information found in the same index for the properties below refers to the keypoint estimated for the nose.
\item \textbf{score}: the score for the keypoint in the respective index, which indicates how confident the estimator is about the accuracy of this keypoint.
\item \textbf{x}, \textbf{y}, and \textbf{z}: the values estimated for the axes \(x\), \(y\), and \(z\) of the keypoint in the respective index.
\end{itemize}

The resulting dataset and the source code used to build it are available through the URL listed below\footnote{Available at \urlDatasetTD}.

\section{\datasetPhono{}}
\label{sec:dataset-phono}

The \datasetPhono{} introduces a novel linguistics-based representation, which describes the signs in the ASLLVD dataset in terms of a set of attributes of the American Sign Language phonology~\cite{brentari2018slphonology}.

It was created by leveraging the three-dimensional keypoints provided for the 9,747 samples in the \datasetTD{} and computing the respective phonological attributes, as described in the sections ahead. The selected attributes consist of:

\begin{itemize}
\item \textbf{Handshape}: shape assumed by both the dominant and non-dominant hands while articulating the signs.
\item \textbf{Orientation}: direction in which each of the palms point in the signs.
\item \textbf{Movement}: movement performed by each of the hands in the signs.
\item \textbf{Mouth opening (non-manual expression)}: opening of the mouth, at each frame, which usually differentiates or provides additional meaning to the signs.
\end{itemize}

\subsection{Handshape}
\label{subsec:handshape}

\begin{figure}
    \centering
    \includegraphics[width=0.9\linewidth]{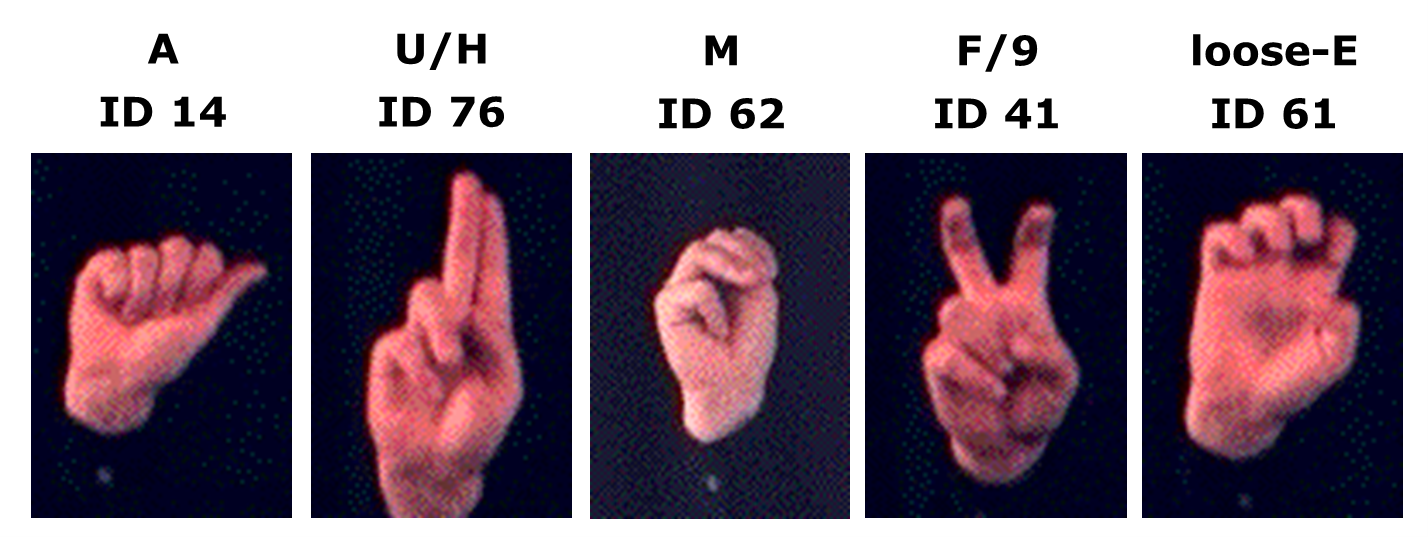}
    \caption{Some handshapes described in the ASLLRP~\cite{neidle2020asllrp}. Here are represented the handshapes "A", "U/H", "M", "F/9", and "loose-E".
    }
    \label{fig:asllrp-handshapes}
\end{figure}

The ASLLVD dataset provides information about the initial and final handshapes for every sample, described according to the 88 shapes presented in the American Sign Language Linguistic Research Project (ASLLRP) (see \autoref{fig:asllrp-handshapes})~\cite{neidle2020asllrp}. 

Thus, we adopted the same information to compute this attribute for the frames of samples in the \datasetPhono{}. However, to assign the proper handshape to each frame in the sequences, we needed to divide them into two halves:
\begin{itemize}
\item The first, containing the frames from the initial part of the sequence, to which we assigned the \textit{initial handshape} from the ASLLVD annotation.
\item The second, with the frames from the final part, to which we assigned the \textit{final handshape}.
\end{itemize}

\subsection{Orientation}
\label{subsec:orientation}

To interpret the orientation of the signers' palms, we resorted to some linear algebra~\cite{anton2013algebra} and explored the relationship between it and the three-dimensional space in which the coordinates are represented.

\begin{figure}
    \centering
    \includegraphics[width=0.45\linewidth]{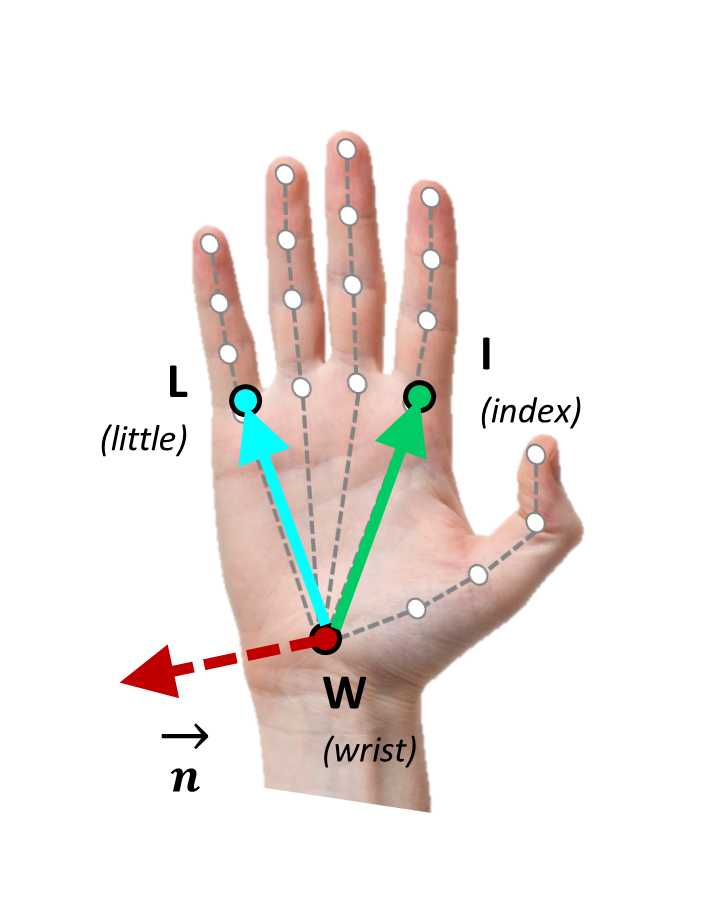}
    \caption{The palm was assumed to be a plane passing through the keypoints estimated for the hand. By selecting the keypoints \(W\), \(L\), and \(I\), it was possible to identify two vectors that provided us with the orientation of the plane and, consequently, of the palm. }
    \label{fig:palm-orientation}
\end{figure}

First, we assumed each palm as a plane that passes through the previously estimated keypoints for the hand joints. Then, from those keypoints, we selected three that could best describe the plane (see the \autoref{fig:palm-orientation}): \(W\) (which refers to the wrist), \(L\) (located at the base of the little finger), and \(I\) (located at the base of the index finger).

Once having these keypoints, we were able to identify two vectors, in terms of which the plane will now be described (see \autoref{fig:palm-orientation}): \(\overrightarrow{WL}\) (the blue arrow) and \(\overrightarrow{WI}\) (the green arrow). Finally, we have the information needed to identify the direction of the palm, which will be done by calculating the normal vector \(\overrightarrow{n}\) (dotted red arrow in \autoref{fig:palm-orientation}), which is perpendicular to the plane:

\begin{equation}
    \label{eqn:normal-palm-left}
    \overrightarrow{n}_{left} = \overrightarrow{WI} \times \overrightarrow{WL}
\end{equation}

\begin{equation}
    \label{eqn:normal-palm-right}
    \overrightarrow{n}_{right} = \overrightarrow{WL} \times \overrightarrow{WI}
\end{equation}

Where \autoref{eqn:normal-palm-left} and \autoref{eqn:normal-palm-right} refer, respectively, to the normal vectors for the left and right palms.

By using \(\overrightarrow{n}\), we classified the palm orientation \(O_{palm}\) as the combination of up to three of the labels described below (for example, \textit{right\_down} or \textit{left\_up\_body}). This was made by evaluating the variations across the axes \(x\), \(y\), and \(z\), as in the equation:

\begin{equation}
\label{eqn:palm-orientation-directions}
     O_{palm} =
         \begin{cases}
              right & \text{if $\overrightarrow{n}_x < {-k}$ }\\
              left  & \text{if $\overrightarrow{n}_x > {k}$ }\\
              up    & \text{if $\overrightarrow{n}_y < {-k}$ }\\
              down  & \text{if $\overrightarrow{n}_y > {k}$ }\\
              body  & \text{if $\overrightarrow{n}_z < {-k}$ }\\
              front & \text{if $\overrightarrow{n}_z > {k}$ }\\
        \end{cases}    
\end{equation}

Where the threshold \(k\) was empirically set to 0.30 (once considered that the coordinates are normalized in the \datasetTD{}) to filter out non-significant variations. See in Figure 8 a visual representation of how \autoref{eqn:palm-orientation-directions} behaves to classify \(\overrightarrow{n}\) relatively to the signer's position in the three-dimensional space.

\begin{figure}
    \centering
    \includegraphics[width=0.7\linewidth]{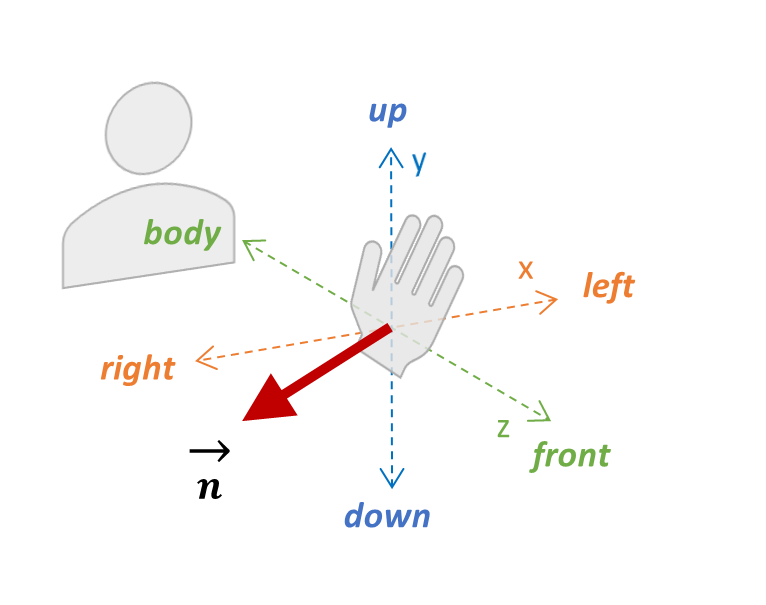}
    \caption{The normal \(\protect \overrightarrow{n}\) was used to determine the palm orientation, which was made by evaluating the values of the axes \(x\), \(y\), and \(z\) relatively to the signer's body. In this example, the palm orientation should be classified as \textit{right\_front\_down}.}
    \label{fig:orientations}
\end{figure}

\subsection{Movement}
\label{subsec:movement}

Similar to what we did for the palm orientation, we also analyzed the estimated keypoints for the hands to determine their trajectory through space.

\begin{figure}
    \centering
    \includegraphics[width=0.6\linewidth]{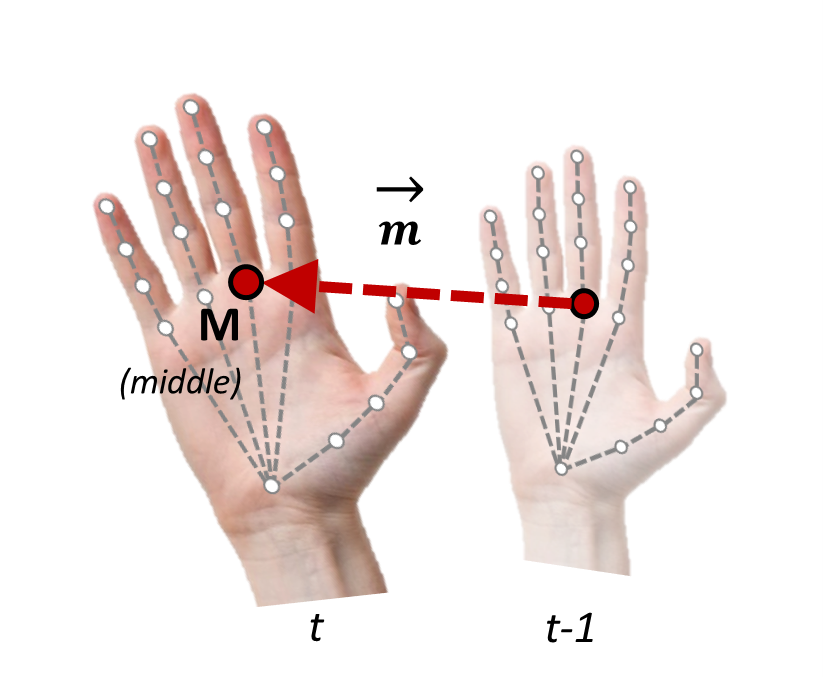}
    \caption{The vector \(\protect \overrightarrow{m}\) was calculated by the trajectory of the keypoint \(M\) between the previous (\(t-1\)) and current (\(t\)) frames.}
    \label{fig:hand-movement}
\end{figure}

For this, we first selected keypoint \(M\) (located at the base of the middle finger) to be used as a reference for the hands (see \autoref{fig:hand-movement}). Then, we calculated its displacement between the previous (time \(t-1\)) and the current (time \(t\)) frames to determine the motion vector \(\overrightarrow{m}\):

\begin{equation}
\label{eqn:hand-movement}
    \overrightarrow{m} = M_{t} - M_{t-1}
\end{equation}

Once \(\overrightarrow{m}\) was calculated, it was possible to classify the hand movement \(M_{hand}\) by using an equation similar to the one used to compute the palm orientation:

\begin{equation}
\label{eqn:hand-movement-directions}
      M_{hand} =
         \begin{cases}
              right & \text{if $\overrightarrow{m}_x < {-k}$ }\\
              left  & \text{if $\overrightarrow{m}_x > {k}$ }\\
              up    & \text{if $\overrightarrow{m}_y < {-k}$ }\\
              down  & \text{if $\overrightarrow{m}_y > {k}$ }\\
              body  & \text{if $\overrightarrow{m}_z < {-k}$ }\\
              front & \text{if $\overrightarrow{m}_z > {k}$ }\\
        \end{cases}    
\end{equation}

\begin{figure}
    \centering
    \includegraphics[width=0.7\linewidth]{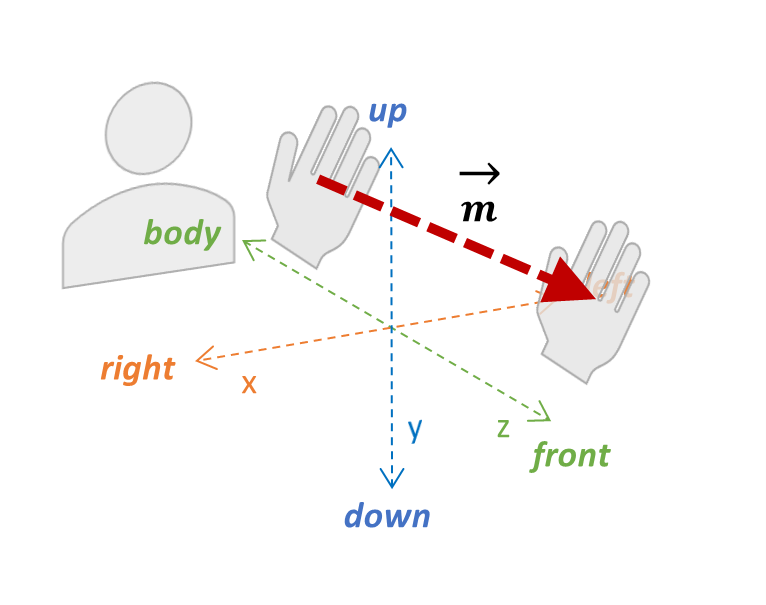}
    \caption{The hand movement is calculated based on the vector \(\protect \overrightarrow{m}\). In this example, we can see a movement classified as \textit{front\_left}, according to the signer's perspective.}
    \label{fig:hand-movement-2}
\end{figure}

Finally, the hand movement is classified as the combination of up to three of the labels in the equation above, and the threshold \(k\) was empirically set to 0.30 to filter out minor movements. \autoref{fig:hand-movement-2} illustrates this classification according to the signer's perspective, within the three-dimensional space.

\subsection{Mouth opening (Non-manual expression)}
\label{subsec:mouth-opening}

For the mouth opening attribute, we considered the research developed by \cite{ferrario2000normal}, which analyzes and establishes some reference parameters for the human lips. 

\begin{figure}
    \centering
    \includegraphics[width=0.50\linewidth]{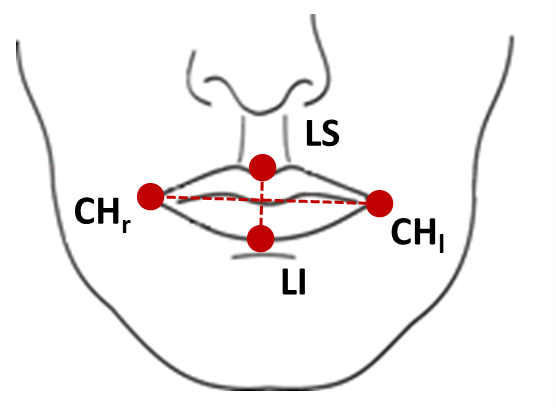}
    \caption{The mouth opening was calculated based on the parameter \textit{vermilion height-to-mouth width} presented by \cite{ferrario2000normal}, which is obtained from the keypoints \(LS\), \(LI\), \(CH_r\), and \(CH_l\).}
    \label{fig:mouth-openness}
\end{figure}

From those parameters, we selected the \textit{vermilion height-to-mouth width} because it can measure the lips in terms of a single ratio (between the vermilion height and the mouth width, as in \autoref{fig:mouth-openness}). Thus, the vermilion height is the distance \(d\) between the labiale superius \(LS\) and labiale inferius \(LI\), which are the outermost keypoints of the upper and lower lips. The mouth width, in turn, is the distance \(d\) between the cheilion right \(CH_r\) and cheilion left \(CH_l\), which are the keypoints on the right and left side of the mouth~\cite{ferrario2000normal}.

This calculation above was adopted to determine the mouth opening \(P_{mouth}\), as follows:

\begin{equation}
    \label{eqn:mouth-openness}
    P_{mouth} = \frac{d(LS, LI)}{d(CH_r, CH_l)}
\end{equation}

\begin{figure}
    \centering
    \includegraphics[width=0.8\linewidth]{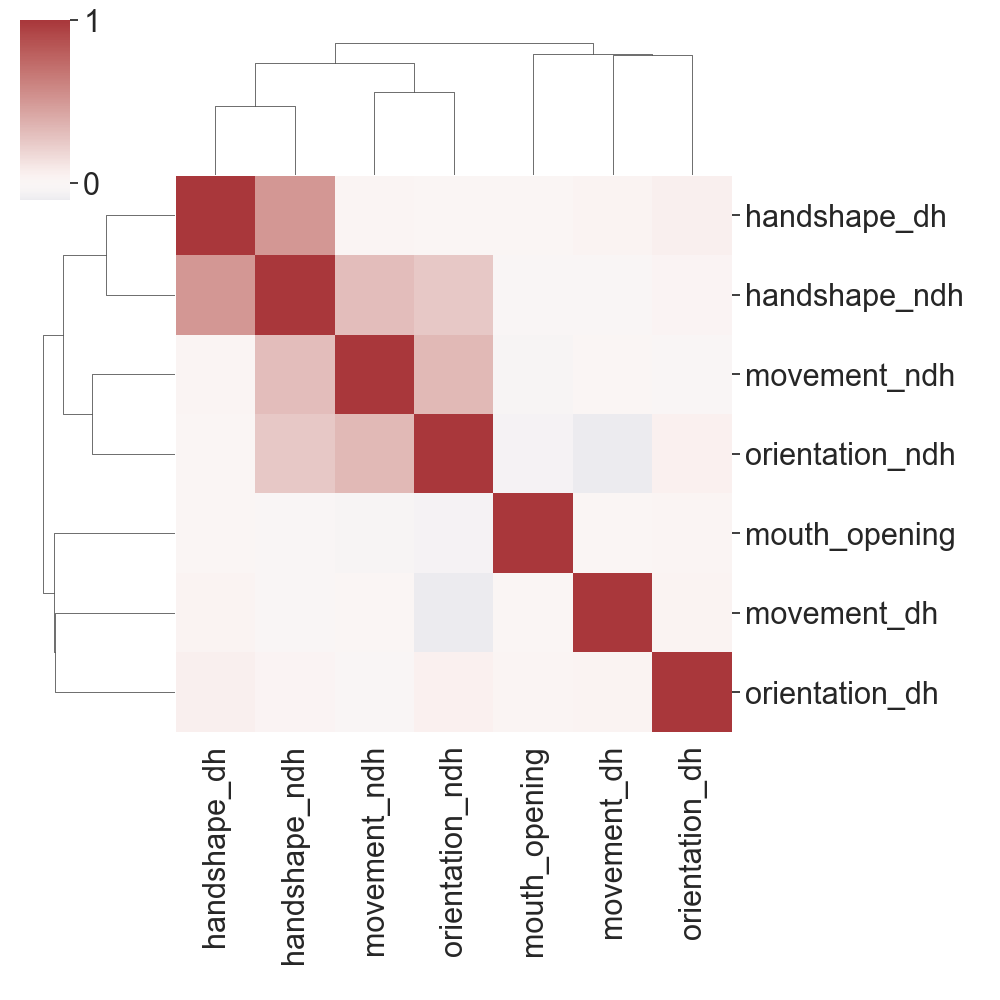}
    \caption{Correlations between the attributes processed for the \datasetPhono{} dataset.}
    \label{fig:attributes-correlation}
\end{figure}

Looking at the set of attributes we just processed in the previous sections, \autoref{fig:attributes-correlation} presents a diagram containing their correlations. It reveals, for example, that the handshapes of the dominant (dh) and non-dominant (ndh) hands are strongly correlated with each other. It is also noted that the handshape of the non-dominant hand significantly correlates with its movement and palm orientation. On the other hand, there are no significant correlations of other attributes with the movement of the dominant hand or its palm orientation. The same is true for the mouth opening, which has no correlations.

In \autoref{tab:dataset-phono-stats} it is possible to observe some statistics calculated from the \datasetPhono{}. The numbers are listed grouped into three perspectives:
\begin{itemize}
\item \textbf{Overall}: considers the total values looking at the dataset as a whole. For example, there are 9,747 distinct samples, 2,284 labels, and 26 movements for the dominant hand.
\item \textbf{Statistics per sample}: calculates values grouping by samples. For example, there is an average of 3.02 frames per sample, with 1 being the minimum and 12 the maximum number per sample. Similarly, the dominant hand has an average of 1.94 distinct movements per sample, with 0 being the minimum and 10 the maximum number.
\item \textbf{Statistics per label}: calculates values grouping by labels. For example, there is an average of 4.10 samples per label, but the ones with a smaller number have at least 2 samples and those with a higher number have up to 59 samples. Similarly, an average of 6.04 distinct movements per label is found for the dominant hand, with this number ranging from 1 to 24.
\end{itemize}

\begin{table*}
    \centering
    
    \begin{threeparttable}
        \caption{Statistics calculated from the \datasetPhono{} dataset.}
        \label{tab:dataset-phono-stats}
    
        \arrayrulecolor{black}
    
        \begin{tabular}{|l|l|rrrrrrrrrr|} 
            \hline
                \rowcolor[rgb]{0.902,0.902,0.902} {\cellcolor[rgb]{0.902,0.902,0.902}} & {\cellcolor[rgb]{0.902,0.902,0.902}} & \multicolumn{1}{c}{{\cellcolor[rgb]{0.902,0.902,0.902}}} & \multicolumn{1}{c}{{\cellcolor[rgb]{0.902,0.902,0.902}}} & \multicolumn{1}{c}{{\cellcolor[rgb]{0.902,0.902,0.902}}} & \multicolumn{2}{c}{\textbf{Movement}} & \multicolumn{2}{c}{\textbf{Orientation}} & \multicolumn{2}{c}{\textbf{Handshape}} & \multicolumn{1}{c|}{{\cellcolor[rgb]{0.902,0.902,0.902}}} \\ 
                \hhline{|>{\arrayrulecolor[rgb]{0.902,0.902,0.902}}----->{\arrayrulecolor{black}}------>{\arrayrulecolor[rgb]{0.902,0.902,0.902}}->{\arrayrulecolor{black}}|}
                \rowcolor[rgb]{0.902,0.902,0.902} \multirow{-2}{*}{{\cellcolor[rgb]{0.902,0.902,0.902}}\textbf{}} & \multirow{-2}{*}{{\cellcolor[rgb]{0.902,0.902,0.902}}\textbf{}} & \multicolumn{1}{c}{\multirow{-2}{*}{{\cellcolor[rgb]{0.902,0.902,0.902}}\textbf{Samples}}} & \multicolumn{1}{c}{\multirow{-2}{*}{{\cellcolor[rgb]{0.902,0.902,0.902}}\textbf{Labels}}} & \multicolumn{1}{c}{\multirow{-2}{*}{{\cellcolor[rgb]{0.902,0.902,0.902}}\textbf{Frames}}} & \multicolumn{1}{c}{DH\tnote{*}} & \multicolumn{1}{c}{NDH\tnote{*}} & \multicolumn{1}{c}{DH} & \multicolumn{1}{c}{NDH} & \multicolumn{1}{c}{DH} & \multicolumn{1}{c}{NDH} & \multicolumn{1}{c|}{\multirow{-2}{*}{{\cellcolor[rgb]{0.902,0.902,0.902}}\begin{tabular}[c]{@{}>{\cellcolor[rgb]{0.902,0.902,0.902}}c@{}}\textbf{Mouth }\\\textbf{opening}\end{tabular}}} \\ 
            \hline
                Overall & Total & 9,747 & 2,284 & - & 26 & 26 & 26 & 26 & 85 & 78 & - \\ 
            \hline
                \multirow{4}{*}{\begin{tabular}[c]{@{}l@{}}Statistics \\per sample\end{tabular}} & Min & - & - & 1 & 0 & 0 & 0 & 0 & 0 & 0 & 0.01 \\
                 & Max & - & - & 12 & 10 & 8 & 6 & 5 & 2 & 2 & 2.19 \\
                 & Mean & - & - & 3.02 & 1.94 & 1.26 & 2.20 & 1.18 & 1.17 & 0.72 & 0.13 \\
                 & Std & - & - & 0.87 & 0.83 & 1.15 & 0.79 & 1.07 & 0.38 & 0.58 & 0.12 \\ 
            \hline
                \multirow{4}{*}{\begin{tabular}[c]{@{}l@{}}Statistics \\per label\end{tabular}} & Min & 2 & - & - & 1 & 0 & 1 & 0 & 1 & 0 & 0.02 \\
                 & Max & 59 & - & - & 24 & 16 & 22 & 14 & 8 & 8 & 0.99 \\
                 & Mean & 4.11 & - & - & 6.04 & 3.66 & 5.37 & 2.63 & 1.81 & 1.17 & 0.13 \\
                 & Std & 2.63 & - & - & 2.93 & 3.18 & 2.29 & 2.35 & 0.99 & 1.14 & 0.07 \\
            \hline
        \end{tabular}
        
        \begin{tablenotes}
          \small
          \item[*] DH stands for Dominant-Hand and NDH stands for Non-Dominant Hand.
        \end{tablenotes}
        
    \end{threeparttable}
\end{table*}

\begin{figure}
    \centering
    \includegraphics[width=0.90\linewidth]{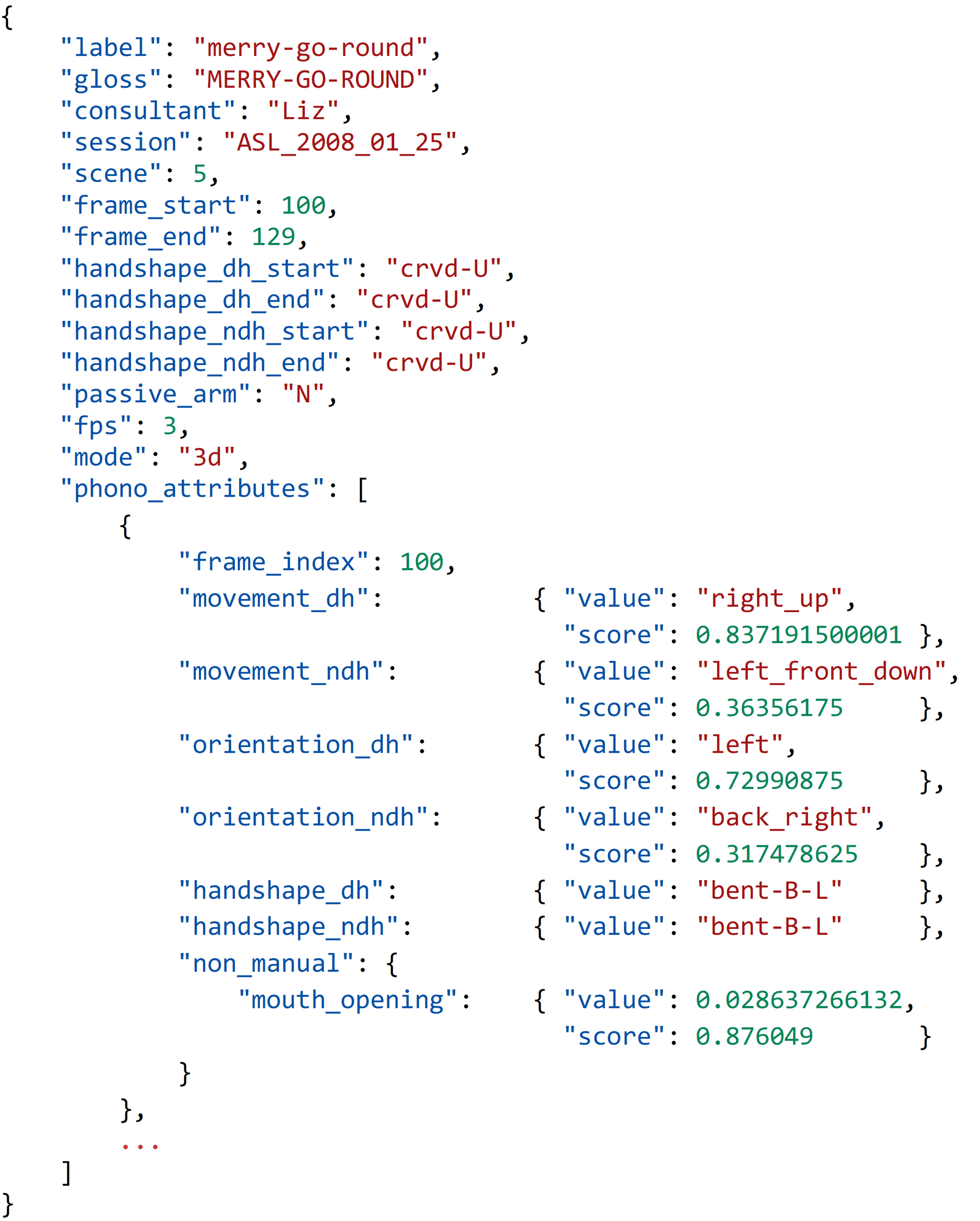}
    \caption{Example of structure and content provided for the samples in the \datasetPhono{} dataset.}
    \label{fig:sample-json-phono}
\end{figure}

\autoref{fig:sample-json-phono} illustrates an example of how the information is structured for the samples in the \datasetPhono{} dataset. The properties at the beginning identify the sign label, session, scene, consultant, and frames in which it was originally recorded. Then, there is a property containing the respective phonological attributes for every frame. Those attributes include the resulting \textbf{value} and a \textbf{score}, which is calculated from the scores of the keypoints involved in the processing.

The resulting dataset and the source code used to build it are available through the URL listed below\footnote{Available at \urlDatasetPhono}.

    \section{Final remarks} 
\label{sec:final-remarks}

This work presented two datasets that aim to contribute as important assets for the American Sign Language, especially helping to overcome the challenges currently encountered by related researches.

The \datasetTD{} provides the three-dimensional coordinates of the signers,  which enables researchers to explore the sign's articulation from a large number of perspectives, as well as simplifies the adoption and analysis of new recognition techniques. Furthermore, including such coordinates instead of videos makes it easier to derive new aggregations and representations that are better suited to specific problem fields.

The \datasetPhono{}, in turn, introduces an unprecedented representation based on the ASL linguistics. Despite considering a limited number of phonological attributes, it allows researchers to study sign language recognition through an entirely new approach, leading to promising findings in this field.

We envision the following items as next steps for the work presented in this paper:

\begin{itemize}
\item Assess the potential gains from adopting the \datasetPhono{} and linguistic-based approaches to SLR, when compared to other traditional datasets and techniques. This is the first step to understand the horizon towards such a novel approach.
\item Evaluate the importance of the phonological attributes selected for the \datasetPhono{} to identify which ones contribute more significantly to SLR tasks and which new ones from the ASL literature should be introduced.
\item From the above discoveries, derive new representations from the \datasetTD{} that allow new approaches and advances for the field of SLR.
\end{itemize}

    \bibliographystyle{IEEEtran}
    \bibliography{IEEEabrv,references.bib}
\end{document}